\documentclass[11pt]{article}

\PassOptionsToPackage{dvipsnames}{xcolor} %

\usepackage[preprint]{acl}

\usepackage{times}
\usepackage{latexsym}
\usepackage{listings}
\usepackage{booktabs}
\usepackage{siunitx}
\usepackage{multirow}
\usepackage{rotating}
\usepackage{placeins}
\usepackage{float}
\usepackage{caption}
\usepackage{subcaption}
\usepackage{tablefootnote}
\usepackage{adjustbox}

\usepackage[most]{tcolorbox}
\tcbuselibrary{skins,breakable}

\definecolor{PromptBlue}{HTML}{3B4BE3} %

\newtcolorbox{promptwindow}[1]{%
  enhanced,
  breakable,
  colback=PromptBlue!6,
  colframe=PromptBlue,
  boxrule=1pt,
  arc=14pt,
  left=14pt,right=14pt,top=14pt,bottom=14pt,
  fonttitle=\bfseries,
  coltitle=white,
  colbacktitle=PromptBlue,
  attach boxed title to top left={xshift=10pt,yshift=-2mm},
  boxed title style={
    sharp corners,
    arc=10pt,
    boxrule=0pt,
    left=10pt,right=10pt,top=6pt,bottom=6pt
  },
  title=#1
}

\newtcolorbox{promptwindowwhite}[1]{%
  enhanced,
  breakable,
  colback=white,
  colframe=PromptBlue,
  boxrule=1pt,
  arc=14pt,
  left=14pt,right=14pt,top=14pt,bottom=14pt,
  fonttitle=\bfseries,
  coltitle=white,
  colbacktitle=PromptBlue,
  attach boxed title to top left={xshift=10pt,yshift=-2mm},
  boxed title style={
    sharp corners,
    arc=10pt,
    boxrule=0pt,
    left=10pt,right=10pt,top=6pt,bottom=6pt
  },
  title=#1
}

\definecolor{NegBiasRed}{HTML}{B24A3A}   %
\definecolor{PosBiasGreen}{HTML}{2E7D5B} %

\definecolor{NegBiasRedTitle}{HTML}{C96A5D}   %
\definecolor{PosBiasGreenTitle}{HTML}{5FAF90} %

\newtcolorbox{biasexample}[3]{%
  enhanced,
  colback=#1!6,
  colframe=#1,
  boxrule=0.8pt,
  arc=8pt,
  left=8pt,right=8pt,top=5pt,bottom=3pt,
  fontupper=\scriptsize,
  fonttitle=\bfseries\footnotesize\color{white},
  colbacktitle=#2,
  coltitle=white,
  attach boxed title to top left={xshift=0pt,yshift=-2mm},
  boxed title style={
    sharp corners,
    arc=6pt,
    boxrule=0pt,
    left=6pt,right=6pt,top=2pt,bottom=2pt
  },
  title=#3
}
\newcommand{\safa}[1]{\textcolor{black}{#1}}

\usepackage[T1]{fontenc}

\usepackage[utf8]{inputenc}

\usepackage{microtype}

\usepackage{inconsolata}
\usepackage{csquotes}

\usepackage{graphicx}
\usepackage{csquotes}

\newcommand{\dataset}{\texttt{LOBSTER}}

\newcommand{\revised}[1]{#1}

 \title{%
     \raisebox{-0.5\height}{\includegraphics[ height=3em]{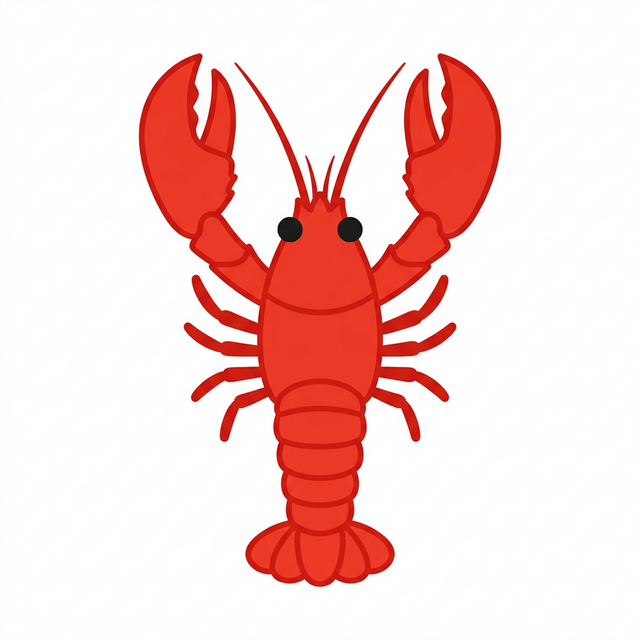}}%
     \hspace{0.5em}%
     \begin{tabular}{c}
         Are Non-English Papers Reviewed Fairly?\\
         Language-of-Study Bias in NLP Peer Reviews
     \end{tabular}%
 }

\author{Ehsan Barkhordar\textsuperscript{1} \qquad Abdulfattah Safa\textsuperscript{1,2} \qquad Verena Blaschke\textsuperscript{3,4}\\ \textbf{Erika Lombart\textsuperscript{5}} \qquad  
\textbf{Marie-Catherine de Marneffe\textsuperscript{5}} \qquad \textbf{Gözde Gül Şahin\textsuperscript{1,2,6}} \\
\textsuperscript{1}Koç University \qquad
\textsuperscript{2} KUIS AI Lab\qquad
\textsuperscript{3}LMU Munich \\ \textsuperscript{4}Munich Center for Machine Learning\qquad
\textsuperscript{5}UCLouvain \\ \textsuperscript{6}Friedrich-Alexander-Universität Erlangen-Nürnberg\\
\url{https://gglab-ku.github.io/}}

\usepackage{graphicx}

\begin{document}
\maketitle
\begin{abstract}
  Peer review plays a central role in the NLP publication process, but is susceptible to various biases. Here, we study language-of-study~(LoS) bias: the tendency for reviewers to evaluate a paper differently based on the language(s) it studies, rather than its scientific merit. Despite being explicitly flagged in reviewing guidelines, such biases are poorly understood. Prior work treats such comments as part of broader categories of weak or unconstructive reviews without defining them as a distinct form of bias. We present the first systematic characterization of LoS bias, distinguishing negative and positive forms, and introduce the human-annotated dataset \dataset{} (Language-Of-study Bias in ScienTific pEer Review) and \revised{an LLM-based detection pipeline achieving 87.37 macro F1}. We analyze 15,645 reviews to estimate how negative and positive biases differ with respect to the LoS, and find that non-English papers face substantially higher bias rates than English-only ones, with negative bias consistently outweighing positive bias. Finally, we identify four subcategories of negative bias, and find that demanding unjustified cross-lingual generalization is the most dominant form. We publicly release all resources to support work on fairer reviewing practices in NLP and beyond.\footnote{\url{https://github.com/GGLAB-KU/LOBSTER}}

\end{abstract}

\section{Introduction}

\begin{figure}[t]
    \centering

    \begin{biasexample}{NegBiasRed}{NegBiasRedTitle}{Negative Bias}
        \textbf{Reviewer comment:}
        \textit{``It would be better if the scope of the paper is larger, e.g., testing the method on Japanese (also an agglutinative language) datasets.''}
        \smallskip

        \textbf{Why this is bias.}
        The paper is explicitly scoped to Korean, and requiring evaluation on another language as a condition for acceptance
        imposes an unjustified multilingual expectation beyond the paper’s stated claims.
    \end{biasexample}

    \begin{biasexample}{PosBiasGreen}{PosBiasGreenTitle}{Positive Bias}
        \textbf{Reviewer comment:}
        \textit{``Under-documented and under-resourced languages should be a priority of the field, and this paper is a valuable contribution.''}
        \smallskip

        \textbf{Why this is bias.}
        The evaluation centers primarily on the choice of language, with little engagement with the paper’s methodology or empirical evidence.
    \end{biasexample}

    \caption{Negative and positive language-of-study bias examples in NLP peer reviews. First review unfairly penalizes a paper for its language-of-study; the second praises it, without  justification, for its language choice.}
    \label{fig:intro-bias-examples}
\end{figure}

Peer review is the cornerstone of scientific evaluation in natural language 
processing (NLP), determining which research gets published. %
While it is meant to ensure quality, peer review is known to be susceptible to various biases such as the gender, writing style, and the affiliation of the authors~\revised{\cite{tomkins2017singleblind,tran2020openreview,lepp2025linguistic}}. Among these, one type of bias has received surprisingly little attention: bias related to the \textit{language a paper chooses to study}. Despite the growth in the number of multilingual studies, NLP has long been centered on English. Research on non-English languages is often treated as niche, and papers are sometimes judged against an implicit English standard rather than on their own stated goals. Furthermore, reviewers sometimes penalize work on non-English languages, calling it ``too narrow,'' or questioning why a particular language was chosen at all. We call this \textit{language-of-study bias}: systematic differences in how a paper is reviewed based on the language(s) it studies, rather than its scientific merit. However bias can also go the other way: some reviewers give unwarranted praise simply because a paper covers a low-resource language, without engaging with the actual methods or results. Figure~\ref{fig:intro-bias-examples} shows two examples from real reviews. 

\revised{The ARR reviewing guidelines explicitly warn against such ``lazy thinking'' heuristics, including dismissing a paper because it is ``tested only on [not English]'' \cite{purkayastha-etal-2025-lazyreview}, and the ACL 2023 report finds that these heuristics account for about 24\% of the problems authors flagged in their reviews \cite{acl2023report}.} Yet, to our knowledge, no prior work systematically defines the types of biases against the language(s)-of-study, and measures how often this bias occurs, in which directions it operates, or what forms it takes.

To address this gap, we present the first large-scale, systematic study of language-of-study bias in NLP peer review. To do so, we first manually investigate a set of review segments carefully sampled from EMNLP 2023, EMNLP 2024 and ACL 2025. %
Then, we construct \dataset{}: Language-Of-study Bias in ScienTific pEer Review, a human-annotated dataset of 529 review segments labeled for negative, positive, or no bias by NLP experts. %
To characterize \textit{where} bias concentrates, we define two auxiliary tasks alongside bias detection: identifying the language(s) studied in each paper and classifying its contribution type (e.g., new model, dataset, or empirical analysis). We benchmark six state-of-the-art LLMs on \dataset{}, and apply the best-performing model to a large corpus of 15,645 reviews spanning six NLP venues, going well beyond the annotated set to estimate bias prevalence at scale. We further perform a qualitative analysis of the negative bias instances in \dataset{}, identifying subcategories that capture the distinct ways language bias manifests in practice. Our analysis reveals that non-English papers face substantially higher bias rates than English-only papers, with negative bias consistently outweighing positive bias, and that the most common pattern is reviewers demanding cross-lingual generalization that was never claimed by the authors.

\section{Related Work}
\label{sec:related-work}

\revised{Prior studies show that reviewer bias is linked to author characteristics such as institutional prestige, location, gender, and writing style. For instance, \citet{tomkins2017singleblind} find that when reviewers know author identities, they favor submissions from established authors and top institutions. Similarly, \citet{tran2020openreview} analyze ICLR submissions on OpenReview and find evidence of gender bias in acceptance decisions. Another study on ICLR reviews finds that reviewers rigorously criticize the writing style of the authors affiliated with institutions in countries where English is less widely spoken \cite{lepp2025linguistic}. These biases operate on \textit{author-side} attributes (identity, affiliation, writing style) that are independent of paper content; which is distinct from our focus on bias against papers \emph{studying} non-English languages, regardless of authorship. We complement this body of work by examining a content-side bias defined relative to each paper's stated scope. Another line of research close to our study is automated analysis of biases in peer reviews. \citet{manzoor2021uncovering} introduce a framework that tests whether reviews are biased against a predefined group of authors, and show that such bias can be detected from the text of reviews alone, without access to review scores. Furthermore, \citet{guo-etal-2023-automatic} use argument mining to check whether the claims in a review are substantiated. Unlike these studies, we exploit state-of-the-art LLMs for automated bias identification and classification.} %

The work closest to ours is LazyReview \cite{purkayastha-etal-2025-lazyreview}, a dataset of review sentences annotated with the ``lazy thinking'' heuristics that the ARR reviewing guidelines caution against. One of these heuristics concerns the language a paper is evaluated on: ``The approach is tested only on [not English], so unclear if it will generalize to other languages''. However, they solely rely on the review text to identify biases, ignoring the paper’s scope and claims. We hypothesize that, this context is crucial. For example, a request for multilingual evaluation may be reasonable if the paper claims generality, but biased if it explicitly focuses on a single language. We therefore frame the problem as detecting bias relative to a paper's stated scope rather than as classifying review sentences in isolation. Furthermore, we distinguish negative bias from positive bias. To the best of our knowledge, this is the first in-depth study on language-of-study~(LoS) bias, providing insights on how negative and positive biases towards the language of study differ with respect to various factors including individual languages, contribution types, and venues.

\section{The \dataset{} Dataset}
\label{sec:datasets}
\begin{table*}[t]
    \centering
    \small
    \scalebox{1}{
    \begin{tabular}{p{1.2cm} p{6.4cm} p{7.2cm}}
        \toprule
        \textbf{Label} & \textbf{Description}                                                                                                                                                                                                                                                                                                                             & \textbf{Example(s)} \\
        \midrule

        \textcolor{NegBiasRed}{\textbf{Negative Bias}}
                       & A reviewer \textcolor{NegBiasRed}{\textbf{devalues}} or \textcolor{NegBiasRed}{\textbf{dismisses}} the research because of the language(s) studied, or assumes a particular language (often English) that is not central to the study is the default, superior, or required standard for demonstrating \textbf{validity} or \textbf{importance}.
                       & \textit{The proposed approach was solely evaluated on three Chinese dialogue datasets. It could be better if the authors would experiment with English dialogue datasets to further demonstrate its effectiveness.} {\scriptsize[\href{https://openreview.net/forum?id=OVt2dIwxR1}{link}]}                                                     %
        \newline
        \textit{The study is very specific to sign language, I don't see how to use it in other tasks.} {\scriptsize[\href{https://openreview.net/forum?id=Ck3JPqoEeE}{link}]}
        \newline
        \textit{that too using a single language (French), which makes me question the applicability and generalizability} {\scriptsize[\href{https://openreview.net/forum?id=V9xsOja2oC}{link}]}
        \\

        \addlinespace

        \textcolor{PosBiasGreen}{\textbf{Positive Bias}}
                       & A reviewer \textcolor{PosBiasGreen}{\textbf{overly praises}} the use of certain languages (e.g., low-resource languages) \textcolor{PosBiasGreen}{\textbf{without engaging}} with methodology, analysis, or contribution.
                       & \textit{More work in low-resource languages is always good.} {\scriptsize[\href{https://openreview.net/forum?id=L8Cxea5krb}{link}]}                                                                                                                                                                                                            %
        \\

        \addlinespace

        \textbf{No Bias Detected}
                       & \textbf{Scientifically grounded comments} aligned with the paper’s scope. Includes \textbf{language related criticism that is relevant to the paper’s stated goals}, as well as comments unrelated to the language(s) studied (e.g., writing quality).
                       & \textit{Limited experimentation with non-English languages.\ (valid since the paper claims multilingual generalization)} {\scriptsize[\href{https://openreview.net/forum?id=2FDty4mLqP}{link}]} \newline
        \textit{How does Bangla word analogy compare with English?} (valid since it is relevant to the paper’s goals) {\scriptsize[\href{https://openreview.net/forum?id=B14ohp9mPU}{link}]} \newline
        \\

        \addlinespace

        \textbf{Needs Context}
                       & Used only when the reviewer’s intent \textbf{cannot be judged} from the paper abstract and review alone.
                       & \textit{Testing on a limited number of languages or training sets is not sufficient to support the claims.}\ (Claims are not given explicitly.) {\scriptsize[\href{https://openreview.net/forum?id=9K1urVN7ti}{link}]}                                                                                                                         \\

        \bottomrule
    \end{tabular}
    }
    \caption{Annotation schema for language-related bias in peer reviews. Review segments are from real reviews.
    }
    \label{tab:bias_schema}
\end{table*}

\safa{We manually investigate peer reviews to identify and define the types of reviewer biases that exist for language(s) studied in a paper. We use three publicly licensed NLP peer-review sources spanning consecutive review cycles: EMNLP~2023, EMNLP~2024, and ACL~2025, totaling 11,680 reviews (Table~\ref{tab:probe-annotation}). Although the sources are review-level, we choose \emph{review segments} extracted from review text (typically a sentence or clause expressing a single critique/stance) as the unit of analysis, to potentially detect multiple bias types in a review.}

\begin{table}[t]
    \centering
    \small
    \setlength{\tabcolsep}{5pt}
    \scalebox{0.82}{
    \begin{tabular}{l rr rr}
        \toprule
                              & \multicolumn{2}{c}{\textbf{Full Corpus}} & \multicolumn{2}{c}{\textbf{\dataset{}}}                                        \\
        \cmidrule(lr){2-3} \cmidrule(lr){4-5}
        \textbf{Venue}        & \textbf{Papers}                          & \textbf{Reviews}                              & \textbf{Papers} & \textbf{Segments} \\
        \midrule
        EMNLP 2023\textsuperscript{$\dagger$}            & 2,020                                    & 6,449                                         & 290             & 375               \\
        EMNLP 2024\textsuperscript{$\dagger$}            & 1,063                                    & 1,425                                         & 99              & 103               \\
        ACL 2025 (Dec--Feb)\textsuperscript{$\ddagger$}   & 2,187                                    & 3,756                                         & 54              & 56                \\
        ARR 2024 (Apr--Jun)\textsuperscript{$\ddagger$}   & 464                                      & 499                                           & --              & --                \\
        COLING/NAACL 2025\textsuperscript{$\ddagger$}     & 410                                      & 498                                           & --              & --                \\
        EMNLP 2025 (Jun--Aug)\textsuperscript{$\ddagger$} & 1,762                                    & 3,018                                         & --              & --                \\
        \midrule
        \textbf{Total}        & \textbf{7,906}                           & \textbf{15,645}                               & \textbf{443}    & \textbf{534}      \\
        \bottomrule
    \end{tabular}
    }
    \caption{Overview of peer-review data used for analysis and annotation. \textbf{Full Corpus}: Review data used for large-scale bias analysis (\S\ref{sec:analysis}). \textbf{\dataset{}}: Annotated subset (\S\ref{ssec:annotation}); each review contributes exactly one segment. Sources: \textsuperscript{$\dagger$}\,NLPEERv2~\cite{dycke2025nlpeer}, \textsuperscript{$\ddagger$}\,ARR Data Collection Initiative~\cite{arrdc-2025-v1.1}.}
    \label{tab:probe-annotation}
    \label{tab:analysis-corpus}
\end{table}

\subsection{Sampling}
\label{subsection:sampling}
Given the large number of reviews, we create a subsample that is likely to contain interesting bias related phenomena for a diverse set of languages. We use a two-stage sampling strategy designed to (i) efficiently surface likely bias cases and (ii) retain coverage of non-biased language mentions.

\paragraph{Stage 1: LLM-assisted \textsc{bias} sampling} The rationale for this stage is the class imbalance. Our initial analysis reveals that language-of-study biases are likely to occur only in a small fraction of review segments. %
Our initial prompt given in Appendix~\ref{app:prompts} relies on heuristically defined bias categories after manual analysis of review segments. Next, we select the segments that are consistently labeled as \textsc{positive bias} or \textsc{negative bias} across models. It should be noted that the models used at this stage are preliminary and result in many False Positive cases. It ensures that the final annotated data is challenging, containing edge cases that can confuse state-of-the-art models.

\paragraph{Stage 2: \textsc{no bias} sampling} To develop models that accurately distinguish between biased and unbiased cases, we sample \textsc{no bias} segments extracted in Stage 1. We group them by (i) the language(s) studied and (ii) the paper's contribution type, to ensure that the final annotated set would not be dominated by a few languages or contribution types. The methods used to detect the language(s) studied and the contribution types are given in \S\ref{sec:methodology}.

\subsection{Bias annotation}
\label{ssec:annotation}
After sampling 534 challenging and balanced set of review segments, we conduct a two-stage human annotation. First, we identified and defined biases towards/against languages studied to write detailed annotation guidelines. Second, every review segment got annotated following the guidelines. Deciding whether a reviewer comment is biased is not straightforward, as it depends on what the paper actually claims. A request for English experiments is unfair when the paper studies Korean morphology and makes no cross-lingual claims, but perfectly valid when the paper explicitly targets multilingual generalization. Therefore, annotators are given full access to the paper (e.g., full PDF, earlier versions) and reviews (e.g., other reviews, rebuttals) through the \texttt{OpenReview} link. These additional data are often consulted for ambiguous cases. \revised{We note that this differs from the LLM inference setup (\S\ref{sec:methodology}), which uses only the paper title, abstract, and \revised{the full review text} \revised{(see Appendix~\ref{app:context-ablation} for details)}}. Each review segment is annotated by at least three annotators,\footnote{During the initial annotation phase, before the team had fully converged on the bias definitions, we assigned five annotators to a subset of segments to better understand sources of disagreement and refine the guidelines.} each of whom has a strong background in NLP and a proven track record in NLP venues. Table~\ref{tab:probe-annotation} summarizes the venue composition of \dataset{}.

\paragraph{First phase.} The first phase served two purposes: (i) to familiarize annotators with the data and the range of language-related comments in reviews, and (ii) to uncover recurring patterns and edge cases that informed the operational definitions of bias labels. We start with the following definition: \textit{Bias is a systematic and unfair deviation from impartial judgment, often caused by irrelevant preferences or assumptions}. We developed the guidelines collaboratively through an iterative process, defining four labels: \textsc{negative bias}, \textsc{positive bias}, \textsc{no bias}, and \textsc{needs context} (Table~\ref{tab:bias_schema}).\footnote{The annotation guidelines are available at \url{https://github.com/GGLAB-KU/LOBSTER/blob/main/annotation_guideline.md}.}

\paragraph{Full-scale annotation.} Once the guidelines were finalized, all annotators revisited their annotations to ensure consistency with the agreed-upon label definitions. %
We obtained substantial inter-annotator agreement \cite{landis1977measurement}: $0.68$ Fleiss $\kappa$~\cite{fleiss1971measuring}. %
Unique gold labels were chosen by majority voting, with adjudication in case of high disagreement. %
We excluded 5 segments with the label \textsc{needs context}. Excluding these ambiguous cases is a limitation of our approach, discussed further in \revised{the Limitations section}. We assess individual annotator accuracy against the final consolidated labels, restricting to annotators who completed at least 90 segments ($n=5$): accuracy ranges between 87.6\% and 97.4\%, and Cohen's $\kappa$~\cite{cohen1960coefficient} ranges between 0.686 and 0.907 (average 0.786), also indicating substantial agreement. \revised{Furthermore, we perform a qualitative analysis on the individual cases with high annotator disagreement in Appendix~\ref{app:annotator-analyses}.}

\paragraph{Final dataset.} Our annotated review dataset contains 529 segments, with 439 \textsc{No Bias}, 73 \textsc{negative bias} and 17 \textsc{positive bias}. Our two-level sampling ensures that \textsc{No Bias} instances contain challenging cases such as language-related comments that are scientifically grounded and justified. Our two-stage annotation ensures high inter-annotator agreement yielding a high quality annotated corpus. Consistent with broader trends in the field, English overwhelmingly dominates the distribution of languages studied, followed by Chinese, Spanish, and French. \dataset{} reflects a diverse range of contribution types, with most papers focusing on NLP applications, new datasets, and analytical studies. Detailed statistics are in Appendix~\ref{app:dataStats}.

\section{Tasks and Experimental Setup}
\label{sec:methodology}
We define three classification tasks: language-of-study detection, contribution type classification and bias classification. While bias classification is the main goal, the two paper-level tasks will help us characterize where bias concentrates. The language-of-study and contribution type tasks are defined as multi-label, bias detection as multi-class.

\subsection{Bias classification}

The goal of this task is to classify a review segment as \textsc{positive}, \textsc{negative} or \textsc{no bias}, given the paper title, abstract and \revised{the full review text}. Due to the small size of the corpus and the challenging nature of the task that requires inferring the claims and scope of the paper for accurate labeling, we exploit six state-of-the-art LLMs (both open source and proprietary): Gemini 3.1 Pro~\cite{google2026gemini31}, Claude Opus 4.6~\cite{anthropic2026claude46}, Grok 4.1~\cite{xai2025grok41}, GPT 5.2~\cite{openai2025gpt52}, DeepSeek V3.2~\cite{deepseekv32pushingfrontieropen}, and Llama 4 Maverick~\cite{meta2025llama4}. %

\begin{table}
    \centering
    \small
    \setlength{\tabcolsep}{4pt}
    \scalebox{0.75}{
    \begin{tabular}{l
            S[table-format=2.2]
            S[table-format=2.2]
            S[table-format=2.2]
            S[table-format=2.2]
            S[table-format=2.2]
            S[table-format=2.2]}
        \toprule
        \textbf{Model}                     &
        \multicolumn{3}{c}{\textbf{Macro}} &
        \multicolumn{3}{c}{\textbf{Weighted}}                                                                                                                \\
        \cmidrule(lr){2-4} \cmidrule(lr){5-7}
                                           & {\textbf{P}} & {\textbf{R}}  & {\textbf{F1}}
                                           & {\textbf{P}} & {\textbf{R}}  & {\textbf{F1}}                                                             \\
        \midrule
        Gemini-3.1-Pro-Preview             & \bfseries 86.47 & \bfseries 88.32 & \bfseries 87.37 & \bfseries 93.63 & \bfseries 93.57 & \bfseries 93.60 \\
        Grok-4.1-Fast                      & 74.34           & 87.81           & 79.75           & 92.43           & 90.36           & 90.96           \\
        GPT-5.2                            & 77.78           & 80.80           & 78.29           & 91.03           & 90.93           & 90.77           \\
        Claude-Opus-4.6                    & 79.68           & 72.45           & 74.96           & 89.23           & 88.85           & 88.91           \\
        DeepSeek-V3.2                      & 61.59           & 80.19           & 66.89           & 87.71           & 79.51           & 81.75           \\
        Llama-4-Maverick-17B               & 59.94           & 74.23           & 63.94           & 85.04           & 76.37           & 79.00           \\
        \midrule
        Random           & 33.36           & 33.36           & 33.33           & 70.85           & 70.88           & 70.85           \\
        Majority         & 27.66           & 33.33           & 30.23           & 68.87           & 82.99           & 75.27           \\     
        \bottomrule
    \end{tabular}
    }
    \caption{Baseline and LLM results on three-way bias classification ($n=529$). Models sorted by Macro F1.}
    \label{tab:model-comparison}
\end{table}

In all experiments, we use the same prompt for each model to ensure fair comparison, refining it iteratively to improve the specification of bias categories and decision rules. Key improvements include clearer distinctions between language-related bias and legitimate methodological critique, explicit handling of edge cases (e.g., multilingual evaluation requests aligned with a paper's scope), and structured output formatting. The complete prompt and inference parameters are provided in Appendix~\ref{app:prompts}.

\begin{figure}[t]
    \centering
    \scalebox{0.7}{
    \includegraphics[width=0.75\linewidth]{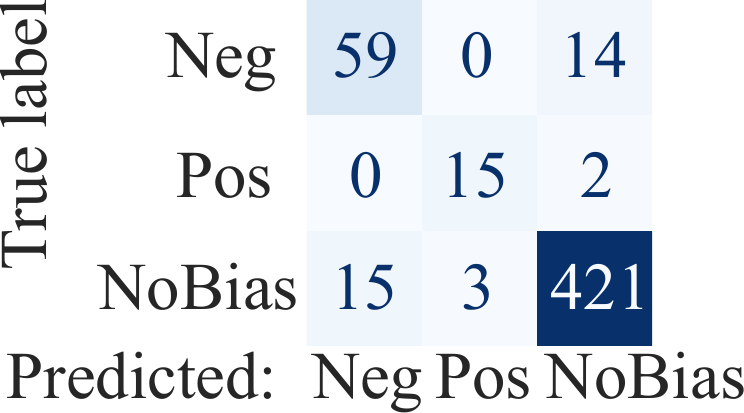}
    }
    \caption{\textsc{Gemini-3.1-Pro-Preview} confusion matrix: \textit{Negative Bias}, \textit{Positive Bias}, \textit{No Bias Detected}.}
    \label{fig:confusion_matrix_multiclass}
\end{figure}

We define two baselines: (i) Majority outputting the most frequent class \textsc{No Bias}, (ii) Random predictions drawing from \dataset{}'s empirical class distribution. We evaluate all models on the full \dataset{} dataset (529 segments, after excluding \textsc{Needs Context}; see \S\ref{ssec:annotation}) using macro- and weighted-F1 scores. Table~\ref{tab:model-comparison} gives the results. All LLMs outperform the baselines by a large margin. \textsc{Gemini-3.1-Pro-Preview} achieves the best Macro F1 and is used in subsequent analyses. False negatives dominate over false positives as shown in Fig.~\ref{fig:confusion_matrix_multiclass}, suggesting a conservative tendency to default to \textsc{No Bias} when language mentions co-occur with legitimate methodological feedback. \revised{We also find that LLM errors tend to fall on segments where annotators disagreed: misclassified segments have higher annotator-vote entropy than correctly classified ones (0.21 vs.\ 0.15), although the difference is not significant on this small set of errors (see Appendix~\ref{app:annotator-analyses}). }%

\subsection{Contribution Type Classification}
We categorize each paper's contribution using a taxonomy grounded in ACL/EMNLP Calls for Papers and ARR reviewing guidelines\revised{, and informed by the contribution taxonomy of \citet{pramanick-etal-2025-nature}}. We define seven categories: Modeling (e.g., novel algorithms), NLP Applications (e.g., pipelines, systems), Data \& Benchmarking, Empirical Analysis, Linguistic Analysis, Domain Adaptation and Survey (see Table~\ref{tab:contribution-categories} in Appendix~\ref{app:dataStats}). %
A paper may receive multiple categories. We manually annotated a sample of 100 papers from three venues (EMNLP 2023, EMNLP 2024, and ACL 2025) to use as an evaluation set. We assign each paper one or more contribution categories based on its title and abstract. We use \textsc{Gemini-3.1-Pro-Preview} with a task-specific prompt given in Appendix~\ref{app:prompts}. The model achieves 82.0\% exact-match accuracy as well as 91.11 Micro F1 and 91.96 Macro F1, demonstrating reliable identification of paper contribution types from metadata alone.  %

\subsection{Language(s) of Study Detection}
To characterize the linguistic scope of each paper\revised{, a task recently benchmarked by \citet{gagnier-kirubakaran-2026-benchmark},} we define a six-category taxonomy (Table~\ref{tab:language-scope-merged}, Appendix~\ref{app:dataStats})
that captures the continuum from single-language studies to language-agnostic work. Papers explicitly listing two or more evaluated languages are labeled \emph{multilingual-specified}; those naming some languages while implying others (e.g., ``English, German, and 8 more'') are \emph{multilingual-partial}; papers providing only a count (e.g., ``101 languages'') are \emph{multilingual-count-only}; and papers making vague claims like ``multilingual'' without any specifics are \emph{multilingual-unspecified}. Work that involves no natural language text (e.g., purely mathematical or symbolic methods) is classified as \emph{language-agnostic}. 
We use macrolanguage names (e.g., Chinese or Arabic)
and include sign languages as distinct natural languages (e.g., American Sign Language). When the abstract does not mention any language but uses known English benchmarks (e.g., SQuAD, GLUE), we default to English. We hand-annotated a sample of  100 papers for the language(s) they study based on their title, abstract and peer reviews, using \textsc{Gemini-3.1-Pro-Preview} with a task-specific prompt (Appendix~\ref{app:prompts}). The model achieves 93\% exact-match accuracy, 95.65 Micro F1 and 83.99 Macro F1. Table~\ref{tab:language-scope-merged} (Appendix~\ref{app:dataStats}) gives per-class precision, recall, and F1 for the language scope categories. Most errors occur on papers that implicitly assume English, or on multilingual papers where the exact set of languages is ambiguous from the available metadata. %

\section{Analysis and Discussion}
\label{sec:analysis}

Using the best-performing LLM from \S\ref{sec:methodology}, we now turn to analyzing biases
in peer reviews across top-tier NLP venues. We apply these models to the
full analysis corpus described in Table~\ref{tab:analysis-corpus} and
address two research questions:
\textbf{RQ1:}~\textit{How do negative and positive biases differ with
respect to the (i)~language(s) studied in the paper and
(ii)~contribution type of the paper?} and
\textbf{RQ2:}~\textit{What subcategories of negative bias exist in
reviews, and how are they distributed across language scopes?}

\subsection{Bias by Language}
\label{subsec:bias-by-language}

We begin by examining how predicted bias distributes across the language(s) studied in a paper.  

\paragraph{English vs.\ non-English.}
First we examine how the bias rates change between the papers that study a single non-English language versus the ones that study English. Table~\ref{tab:bias-by-language-major} reveals a clear divide. Reviews
of English-focused papers exhibit a bias rate of 0.37\%, while reviews of single non-English papers show a
collective rate of 14.79\%, roughly 40 times higher.
This gap persists across multilingual categories: specified multilingual
papers show a rate of 4.18\%, while unspecified
multilingual (0.34\%) and language-agnostic (0.30\%) papers fall close to
the English baseline. The more visibly a paper focuses on non-English
languages, the more likely its reviews are to contain bias. 
The near-zero rate for unspecified multilingual papers is noteworthy.
It may partly reflect a limitation of our classifier, which cannot
attribute bias when specific languages appear only in the body text.
However, it may also reflect a genuine effect: papers that use the
label ``multilingual'' without naming specific languages may preempt the
most common form of bias (generalizability demand) while avoiding
exposure of individual language choices to scrutiny.

\paragraph{Variation for non-English languages.}
Within the non-English category (Table~\ref{tab:bias-by-language-major}),
rates range from 25\%
(Bengali, Greek) to 10.51\% (Chinese). However, most individual
languages have fewer than 50 reviews, making their rates unreliable as
point estimates. Chinese is the exception: with 352 reviews, it provides
the most stable estimate at 10.51\%, more than 28 times the English
rate. Beyond Chinese, we focus on the aggregate finding: non-English
languages as a group show consistently higher bias than English.

\begin{table}[t]
    \centering
    \small
    \setlength{\tabcolsep}{5pt}
    \adjustbox{max width=\linewidth}{%
    \begin{tabular}{@{}l@{}rrrrr@{\,}r@{}}
        \toprule
        \textbf{Language}        & \llap{\textbf{Papers}} & \llap{\textbf{Reviews}} & \llap{\textbf{Bias}} & \textbf{\%} & \textbf{Neg} & \textbf{Pos} \\
        \midrule
        Bengali                  & 11              & 24               & 6            & \textbf{25.0}               & 3            & 3     \\
        Greek                    & 6               & 12               & 3  & 25.0                        & 2            & 1               \\
        Arabic                   & 27              & 47               & 10 & 21.3             & 6            & 4                          \\
        French                   & 8               & 20               & 4     & 20.0           & 4            & 0                        \\
        German                   & 18              & 36               & 7   & 19.4             & 7            & 0                        \\
        Korean                   & 16              & 31               & 5  & 16.1             & 4            & 1                         \\
        Japanese                 & 14              & 28               & 3     & 10.7           & 3            & 0                        \\
        Chinese                  & 178             & 352              & 37    & 10.5           & 31           & 6                        \\
        \midrule
        \textit{Total Single Non-Eng\rlap{.}}      & \textit{350}             & \textit{676}              & \textit{100}      & \textit{14.8}             & \textit{71}           & \textit{31}                   \\
        \textit{English}                  & \textit{6,152}           & \textit{12,125}           & \textit{45}  & \textit{0.4}              & \textit{43}           & \textit{2}                         \\
        \midrule
        Specified Multi.   & 903             & 1,913            & 80          & 4.2        & 51           & 31                      \\

        Unspecified Multi. & 333             & 595              & 2            & 0.3             & 1            & 1               \\
        Language Agnostic        & 168             & 336              & 1       & 0.3           & 1            & 0                      \\
        \bottomrule
    \end{tabular}
    }
    \caption{Model-predicted language-of-study bias rates for single-language papers with at least 5 papers and 10 reviews, plus aggregate language-scope categories. Specified Multilingual denotes papers that name target languages; Unspecified Multilingual denotes count-only or vague multilingual cases. \textbf{Bias}: Absolute and relative number of review segments with any predicted bias. \textbf{Neg/Pos}: Negative/positive bias count.\protect\footnotemark}
    \label{tab:bias-by-language-major}
\end{table}
\footnotetext{A single review may contain both negative and positive bias segments. This did not occur in the manually annotated \dataset{} but arose in 4 of 15,645 reviews in the model-predicted corpus, causing Neg~+~Pos to slightly exceed Biased for some rows.}

\paragraph{Negative vs.\ positive bias.}

\begin{figure}[t]
    \centering
    \includegraphics[width=\columnwidth]{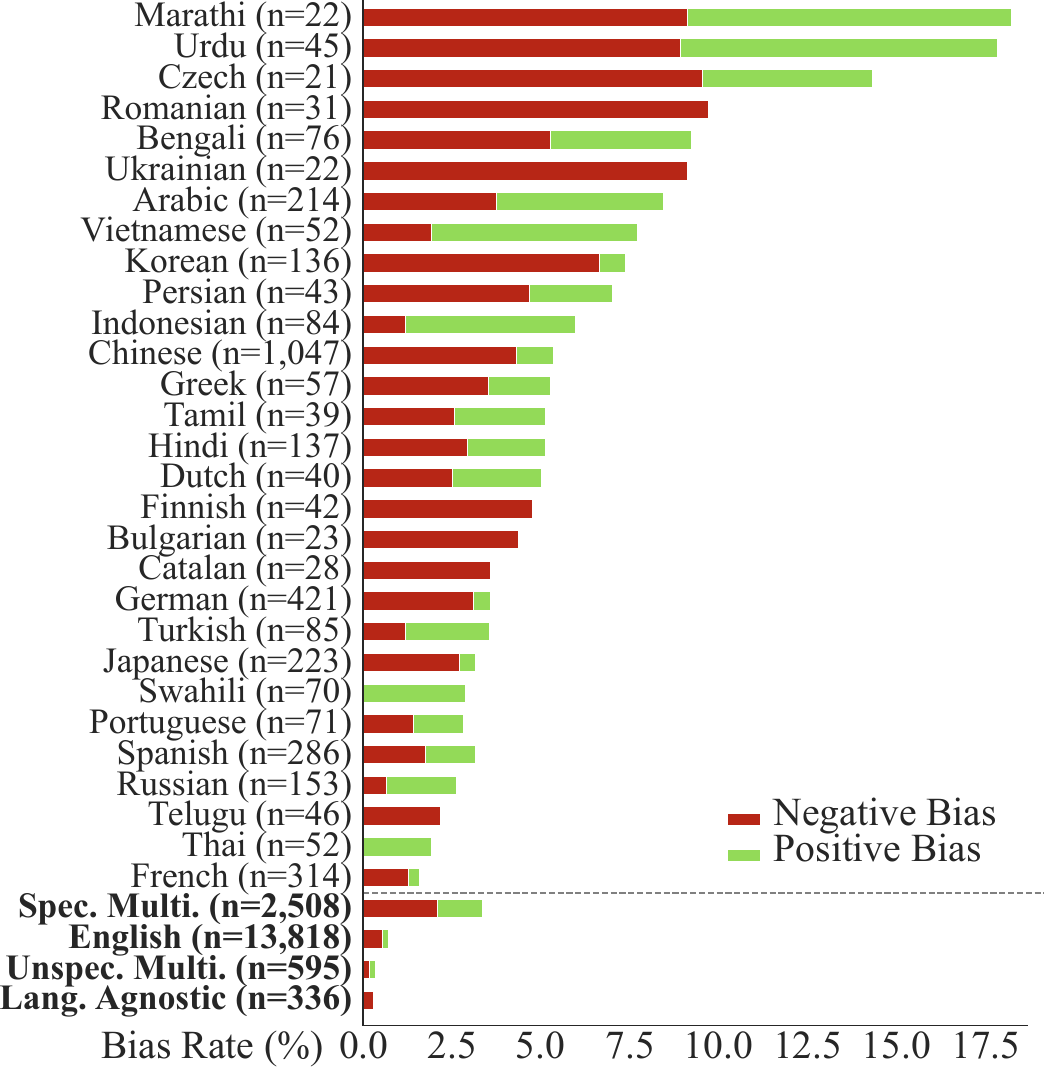}
    \caption{Bias rate (x-axis, \%) by polarity across languages with at least 20 reviews; $n$ denotes the number of reviews.
    \textbf{Top:}  Bias rates for papers mentioning  languages, both single-language and multilingual papers. 
    \textbf{Bottom:} Specified multilingual includes \textit{multilingual-specified} and \textit{multilingual-partial}, while unspecified multilingual contains \textit{multilingual-unspecified} and \textit{multilingual-count-only}.
}
    \label{fig:bias-decomposition-neg-pos-table7}
\end{figure}

In Table~\ref{tab:bias-by-language-major}, papers on a single language other than English
show 71 reviews with negative bias and 31 with positive, a ratio of
about 2.3:1. For Chinese, the imbalance is steeper: 31 negative versus
6 positive (roughly 5:1), indicating that reviewers penalize non-English
language focus far more often than they reward it. English papers show an even stronger skew (43 negative vs.\ 2 positive), but at a much lower base rate.\footnote{\revised{The 2 English positive-bias instances involve papers on English \emph{dialects}, where reviewers praised studying non-standard varieties (e.g., ``Dialects are, in my opinion, under-researched'') rather than engaging with methodology, a pattern analogous to positive bias observed for non-English languages.}} Specified multilingual papers present a different picture. Their
negative-to-positive ratio is about 1.6:1, meaning positive bias
accounts for nearly 39\% of biased reviews in this category, compared to
31\% for single non-English papers and just 4\% for English. Reviewers may see multilingual coverage as a strength by itself, which can bias evaluation toward the choice of languages rather than the quality of the method. To expand our analysis to more languages, we provide a more detailed per-language view of the polarity breakdown in Figure~\ref{fig:bias-decomposition-neg-pos-table7}. Here, a negatively biased review segment for a multilingual paper that studies the languages A and B, counts towards negative biases separately for language A and B. Here, Chinese, for example, shows a roughly 4:1 ratio of negative to positive bias%
across all papers mentioning Chinese (including multilingual ones, hence higher than the single-language counts in Table~\ref{tab:bias-by-language-major}), confirming the negative-dominant pattern. The
overall takeaway is that language-of-study bias is largely
negative: reviewers are more likely to penalize a paper for studying a
non-English language than to credit it for doing so. However, we also observe a noisy yet recurring pattern of pronounced positive biases for several low-resource languages, including Marathi, Vietnamese, Indonesian, and Swahili. Since the number of reviews for these languages is small (see sample sizes in Figure~\ref{fig:bias-decomposition-neg-pos-table7}), we note this as a trend that warrants further investigation rather than a firm conclusion.
We note that the negative-to-positive ratio across the full corpus (2.6:1) is lower than in \dataset{} (4.3:1). This difference traces to two factors. First, the Stage~1 LLM triage (\S\ref{subsection:sampling}) flagged 299 negative candidates but only 38 positive ones (7.9:1), because negative bias patterns (e.g., demands for English evaluation) are more lexically salient and easier for the preliminary models to detect than the subtler positive patterns (e.g., ungrounded praise for language choice). This skewed the annotated pool toward negative cases. Second, the full corpus covers six venues with a larger share of specified multilingual papers, whose negative-to-positive ratio is nearly balanced (1.6:1); these papers pull the corpus-wide ratio well below the \dataset{} ratio.

\subsection{Bias by Contribution Type}
\label{subsec:bias-by-contribution}

Table~\ref{tab:bias-type-by-contrib} shows how bias rates vary across
contribution types. Data \& Benchmarking exhibits the highest negative
bias rate (2.19\%), while Modeling shows the lowest (0.66\%).
This gradient aligns with how visible language choice is in each
contribution type. Data \& Benchmarking work is tied to specific
languages by design: a dataset is built \emph{for} a language. The same holds for Linguistic
Analysis,\footnote{Linguistic Analysis has a smaller sample than the other categories, so its rate should be treated with caution.} where the language studied is also central to the research question.
Methods and Modeling papers, by contrast, tend to frame their
contributions as language-agnostic, reducing the chance that reviewers
engage with language choice at all. This complements the finding in
\S\ref{subsec:bias-by-language}: bias is highest when language choice
is most visible, either due to the language itself or the nature of the
contribution.
We also examine bias rates across venue years in Appendix~\ref{app:bias-figures} (Figure~\ref{fig:bias-type-by-year}); rates appear slightly lower in later cycles, but the time span is too short and venue composition varies too much across years to draw firm temporal conclusions.

\begin{table}[t]
    \centering
    \small
    \scalebox{0.98}{
    \begin{tabular}{lrrrr}
        \toprule
        \textbf{Contribution Type} & \textbf{Reviews} & \textbf{Neg (\%)} & \textbf{Pos (\%)} \\
        \midrule
        Data \& Benchmarking       & 4,423            & \textbf{2.19}     & \textbf{1.06}     \\
        Linguistic Analysis        & 286              & 2.10              & 1.05              \\
        Domain Adaptation         & 1,862            & 1.77              & 0.38              \\
        Empirical Analysis         & 3,567            & 1.21              & 0.36              \\
        NLP Applications         & 7,613            & 0.80              & 0.33              \\
        Modeling                   & 4,544            & 0.66              & 0.11              \\
        \bottomrule
    \end{tabular}
    }
    \caption{Review-level negative and positive bias rates by contribution type. Survey/Position and Other categories are excluded as they show zero bias instances.}
    \label{tab:bias-type-by-contrib}
\end{table}

\subsection{Negative Bias Subcategories}
\label{subsec:bias-subcategories}

To better understand the nature of language-of-study bias, we conducted a manual
qualitative analysis of the 73 instances labeled as \textit{Negative Bias}\footnote{There are not enough cases of positive bias for analysis.} in \dataset{}. We identified four recurring patterns through
which language bias manifests in peer reviews, summarized in Table~\ref{tab:negative-bias-patterns}. \textbf{A-Generalizability Demand} occurs where reviewers penalize papers for not demonstrating cross-lingual generalizability that was never
claimed, treating multilingual coverage as an unstated precondition for methodological
validity. This pattern reflects a common implicit expectation that NLP contributions
must generalize across languages to be considered complete, regardless of whether the
paper's scope is explicitly and deliberately limited to a single language. \textbf{B-English as the Gold Standard} occurs when reviewers
explicitly name English as the missing validation standard, framing non-English results
as insufficiently credible without English corroboration. Unlike \textbf{A}, which demands
more languages in the abstract, this pattern is directional: English is positioned as the
required reference against which non-English contributions must be measured. \textbf{C-Language Choice Interrogation} captures cases where reviewers
question the motivation for studying a particular language, asking for justification
for selecting it, a standard that would not be applied to papers studying English
or other high-resource languages. \textbf{D-Dismissing Impact} occurs where reviewers accept the paper's validity but
minimize the impact, either by arguing that the community served by the paper is too
small, or by denying that studying a particular language constitutes genuine novelty.
This pattern is particularly hard to detect because it can coexist with genuine praise for the
paper's content, with the bias operating at the level of audience size rather than
scientific methodology.

\begin{figure}[t]
    \centering
    \includegraphics[width=\columnwidth]{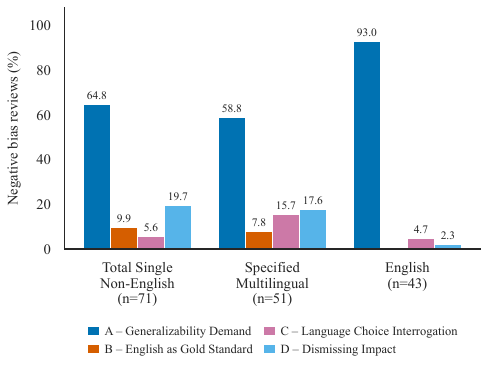}
    \caption{Distribution of negative bias subcategories (A--D) across language scope, predicted by the validated classifier on the full corpus.}
    \label{fig:subcategory-by-language-scope}
\end{figure}

\begin{table*}[t]
    \centering
    \small
    \setlength{\tabcolsep}{4pt}
    \scalebox{1}{
    \begin{tabular*}{\textwidth}{p{4.5cm} S[table-format=2.1] p{10cm}}
        \toprule
        \textbf{Pattern} & \textbf{\%} & \textbf{Example Reviewer Quote} \\
        \midrule
        A-Generalizability Demand &
        62.16 &
        \textit{The proposed method is only tested in Chinese, not for other unsegmented languages.} {\scriptsize[\href{https://openreview.net/forum?id=RXIYmRUWGD&noteId=V8bWBOjWjQ}{link}]} \\[6pt]
        B-English as the Gold Standard &
        9.46 &
        \textit{English has a broader applicability, and the authors could also consider incorporating English annotations.} {\scriptsize[\href{https://openreview.net/forum?id=NO5dc8Ljvj&noteId=0GKSZYu6FS}{link}]} \\[6pt]
        C-Language Choice Interrogation &
        12.16 &
        \textit{Why restricting it to Sanskrit? There are many languages that exhibit productive compounding (e.g.\ German).} {\scriptsize[\href{https://openreview.net/forum?id=8Rif7M7Z6A&noteId=3JQDdB9PWZ}{link}]} \\[6pt]
        D-Dismissing Impact &
        16.22 &
        \textit{The study is narrowly focused on Ancient Greek, which limits its generalizability to other historical or low-resource languages.} {\scriptsize[ACL-ARR-2025]} \\[6pt]
        \bottomrule
    \end{tabular*}
    }
    \caption{Negative language bias patterns and proportion (\%) in the 73
        instances labeled as \textit{Negative Bias} in \dataset{}.}
    \label{tab:negative-bias-patterns}
\end{table*}

\paragraph{Distribution in \dataset{}.}
\textbf{A}-\revised{Generalizability} demand is the most common pattern (62.16\%), which
shows how widespread the implicit expectation of cross-lingual generalizability is among
reviewers, even when the paper never claimed it. \textbf{D}-Dismissing impact comes second (16.22\%) and
is hard to spot because it often appears alongside genuine praise: the bias
shows up in how the audience is valued, not in how the science is judged. \textbf{C}-Language Choice
(12.16\%) is a double standard: reviewers ask authors to justify their language choice,
something they would never ask of an English-focused paper. Finally, \textbf{B}-English as Standard is the
least common (9.46\%); unlike \textbf{A}, which vaguely demands more languages, \textbf{B}
is directional: English is explicitly named as the benchmark that non-English work must
be measured against.

\paragraph{Classifier validation.}
To scale the analysis beyond manual annotation, we evaluated \textsc{Gemini 3.1 Pro} as a
zero-shot classifier on this four-class classification task. \textbf{A}-Generalizability Demand and \textbf{C}-Language Choice
were classified most reliably (97\% and 96\% F1, respectively), while \textbf{D}-Dismissing Impact and
\textbf{B}-English as Standard are more challenging (86\% and 89\% F1). %
Error analysis reveals confusion at two boundaries: between \textbf{C} and \textbf{A} (a language-choice interrogation misclassified as a \revised{generalizability} demand) ; and between \textbf{D} and \textbf{B} (impact dismissal misclassified as English-as-standard). In each case, a single reviewer quote triggers competing pattern signals, making disambiguation difficult even for human annotators.

\paragraph{Corpus-level distribution by language scope.}
Using the validated subcategory classifier, we predict patterns across
all negatively biased reviews in the full corpus.
Figure~\ref{fig:subcategory-by-language-scope} reveals that the mix of
patterns shifts substantially with language scope. For English-only
papers (43 instances), Pattern~\textbf{A} accounts for 93\%, with only marginal
\textbf{C} (4.7\%) and \textbf{D} (2.3\%). Non-English single-language papers (69
instances) show a different profile: \textbf{A} drops to 65.2\%, while \textbf{D} rises to
18.8\% and \textbf{B} to 10.1\%. Specified multilingual papers (49 instances) are
the most diverse: \textbf{A} at 59.2\%, \textbf{C} and \textbf{D} tied at 16.3\%, and \textbf{B} at 8.2\%.
This progression reinforces the findings from
\S\ref{subsec:bias-by-language}: not only do non-English papers attract
more bias overall, but the bias they face is also more varied. English
papers primarily encounter a single pattern (Generalizability Demand),
while non-English and multilingual papers face a wider range of
challenges, including dismissal of their impact and interrogation of
their language choice.

\section{Conclusion}
\label{sec:conclusion}

In this work, we provide the first systematic evidence that language-of-study bias in NLP peer review is not an isolated artifact but a structural pattern. Our large-scale analysis shows that non-English papers face bias rates roughly 40 times higher than English-only ones, consistently across all six venues we examined. Importantly, the problem compounds: while English papers encounter primarily one bias pattern (generalizability demands), non-English and multilingual papers face a wider repertoire, including English-as-gold-standard framing, language choice interrogation, and impact dismissal. These findings point to concrete interventions: reviewer guidelines could explicitly require evaluation against a paper's stated scope, and the strong performance of our LLM-based detector (87.37 Macro F1) suggests automated screening is a realistic complement to human oversight. We release \dataset{}, our annotation guidelines, and the detection pipeline as a foundation for developing fairer reviewing practices.

\section*{Limitations}

\revised{Our study has several limitations. First, although annotating full reviews provides complete context, we adopt conservative labeling in ambiguous or borderline cases, which may ultimately undercount instances of subtle bias. Trained annotators are also hard to find, which limits the size of the dataset. Second, five segments (0.9\%) labeled \textsc{needs context} are excluded from evaluation. Third, our data comes from a limited set of venues and cycles, so prevalence estimates may not generalize to other review systems. Moreover, public review data comes mostly from accepted papers; if biased reviews push papers toward rejection, our estimates likely understate the true prevalence of bias. Fourth, distinguishing language-related bias from valid language-scoped critique is inherently difficult, and borderline cases can yield annotator disagreement. Fifth, we examine language scope, contribution type, venue, and year as independent dimensions, but these factors are likely correlated. For example, non-English papers may cluster in contribution types that are themselves more bias-prone, so differences observed along one dimension may partly reflect variation in another. Fully separating these effects would require a more controlled analysis, which we leave for future work. Finally, our results reflect a particular LLM and prompt version; performance may vary under different configurations. We note that proprietary serving stacks are not strictly deterministic even with a fixed temperature, top-$p$, and seed (e.g., due to mixture-of-experts routing and internal serving configuration), so absolute scores vary slightly across runs. All comparisons we report are made within a single run.}

\section*{Ethical Considerations}
\paragraph{Data licensing.} All review data used in this study come from openly licensed sources: NLPEERv2~\cite{dycke2025nlpeer} and the ARR Data Collection Initiative~\cite{arrdc-2025-v1.1}, both of which were collected with explicit author consent. We release the \dataset{} dataset under a CC~BY~4.0 license to facilitate reproducibility while requiring attribution.

\paragraph{Annotator background and compensation.} Annotation was carried out by the paper's co-authors together with two additional researchers acknowledged below. All annotators hold graduate-level expertise in NLP and participated voluntarily as part of their research activities; no annotators were separately compensated for this work. All annotators were informed of the study's goals and consented to contributing.

\paragraph{Privacy.} Our analysis operates on publicly released review text and paper metadata. We do not attempt to identify individual reviewers, and all review excerpts reproduced in this paper are drawn from publicly accessible OpenReview pages.

\paragraph{Use Of AI Assistants} Claude Opus was used for code development assistance, making the codebase easier and faster to develop. We also use AI assistants for spelling correction, and grammar checks to improve writing.

\section*{Acknowledgments}
\label{sec:acknowledgments}
We thank Dr.\ Ali H\"urriyeto\u{g}lu and Tamta Kapanadze for their help with the annotation effort, including participation in adjudication discussions and guideline refinement.
\revised{Part of this work was initiated by Dagstuhl Seminar \href{https://www.dagstuhl.de/25301}{25301} ``Linguistics and language models: What can they learn from each other?''.}
Marie-Catherine de Marneffe is a Research Associate of the Fonds de la Recherche Scientifique – FNRS.
\revised{Verena Blaschke's research was supported by European Research Council (ERC) Consolidator Grant DIALECT 101043235.
This work has been partially supported by the Scientific and Technological Research Council of Türkiye~(TÜBİTAK) as part of the project ``Automatic Learning of Procedural Language from Natural Language Instructions for Intelligent Assistance'' with the number 121C132. We also gratefully acknowledge KUIS AI Lab for providing computational support.} Finally, this research is with support from Google.org and the Google Cloud Research Credits program for the Gemini Academic Program.

\bibliography{custom}

\clearpage

\appendix
\label{sec:appendix}
\section{Extraction and Normalization Details}
\label{app:extraction-normalization}
To standardize reviewer text across sources, we apply the following normalization steps: (i)~normalize whitespace and unicode quirks (e.g., non-breaking spaces), (ii)~drop empty or near-empty text fields, and (iii)~preserve the original wording and punctuation of reviewer text (no rewriting), since subtle lexical choices are central to bias analysis.

\section{Classification Definitions and Evaluation}
\label{app:dataStats}

Table~\ref{tab:language-scope-merged} defines the six language scope categories and reports per-class precision, recall and F1 on the 100 hand-annotated papers. Table~\ref{tab:contribution-categories} defines the seven contribution type categories.

\begin{table*}[ht]
  \centering
  \small
  \begin{tabular*}{\textwidth}{@{\extracolsep{\fill}} p{4.5cm} p{7cm} c c c c}
    \toprule
    \textbf{Language Scope}  & \textbf{Description}                             & \textbf{P} & \textbf{R} & \textbf{F1} & \textbf{N} \\
    \midrule
    Single-language          & One specific language studied                    & 0.96       & 0.99       & 0.97        & 86 \\
    Multilingual-specified   & Multiple specific languages listed               & 0.75       & 1.00       & 0.86        & 3  \\
    Multilingual-partial     & Some languages named, others implied             & 1.00       & 0.67       & 0.80        & 3  \\
    Multilingual-count-only  & Only a count given (e.g., ``101 languages'')     & --         & --         & --          & -- \\
    Multilingual-unspecified & Vague ``multilingual'' claim, no names or counts & 0.80       & 0.67       & 0.73        & 6  \\
    Language-agnostic        & No natural language involved                     & 0.00       & 0.00       & 0.00        & 2  \\
    \midrule
    Accuracy                 &                                                  & \multicolumn{3}{c}{0.94}              & 100 \\
    \bottomrule
  \end{tabular*}
  \caption{Language scope categories alongside per-class evaluation metrics on 100 hand-annotated papers. P = Precision, R = Recall, N = support. Both language-agnostic samples were misclassified as single-language (English).}
  \label{tab:language-scope-merged}
\end{table*}

\begin{table*}[ht]
  \centering
  \small
  \begin{tabular*}{\textwidth}{@{\extracolsep{\fill}} p{4.5cm} p{10cm}}
    \toprule
    \textbf{Category}    & \textbf{Description}                                                               \\
    \midrule
    Modeling             & New architecture, learning algorithm, or objective function                        \\
    NLP Applications   & Pipeline, prompting strategy, or system built on existing models                   \\
    Data \& Benchmarking & New dataset, corpus, benchmark, or annotation resource                             \\
    Empirical Analysis   & Comparative evaluation, ablation, or interpretability study                        \\
    Linguistic Analysis  & Computational study of language structure, variation, or human language processing \\
    Domain Adaptation   & NLP applied to a specific domain (e.g., clinical, legal)                           \\
    Survey / Position    & Literature review, meta-analysis, or position paper                                \\
    \bottomrule
  \end{tabular*}
  \caption{Contribution type categories used to classify papers.}
  \label{tab:contribution-categories}
\end{table*}

\begin{figure*}
  \centering
  \includegraphics[width=0.75\linewidth]{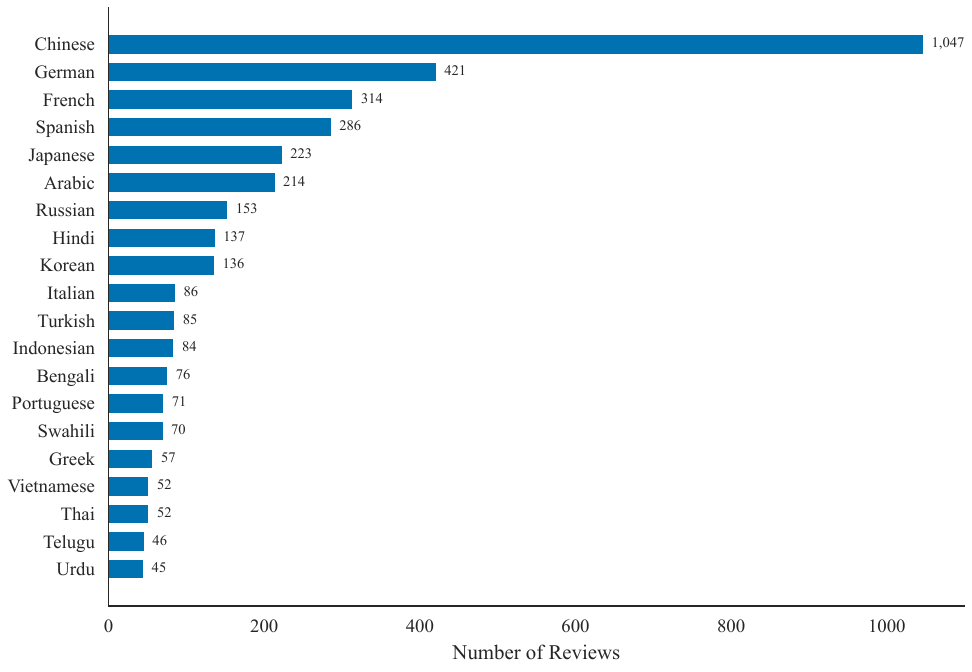}
  \caption{Top 20 non-English studied languages in the full analysis corpus (Table~\ref{tab:analysis-corpus}), inferred from paper titles and abstracts. Counts reflect the number of reviews associated with each language; a paper may study multiple languages and each review is counted for all its languages. English ($n{=}6{,}946$ reviews) is excluded due to its dominant frequency.}
  \label{fig:top-languages}
\end{figure*}

Figure~\ref{fig:top-languages} shows the distribution of the top 20 non-English studied languages in our full analysis corpus, as inferred from paper titles and abstracts using the language identification prompt (see Appendix~\ref{app:prompts}). Counts reflect the number of reviews associated with each language; a single paper may study multiple languages, and each review is counted for all its languages. English is excluded due to its dominant count. Chinese leads among the remaining languages, followed by German and other typologically diverse languages.

\begin{figure*}
  \centering
  \includegraphics[width=0.75\linewidth]{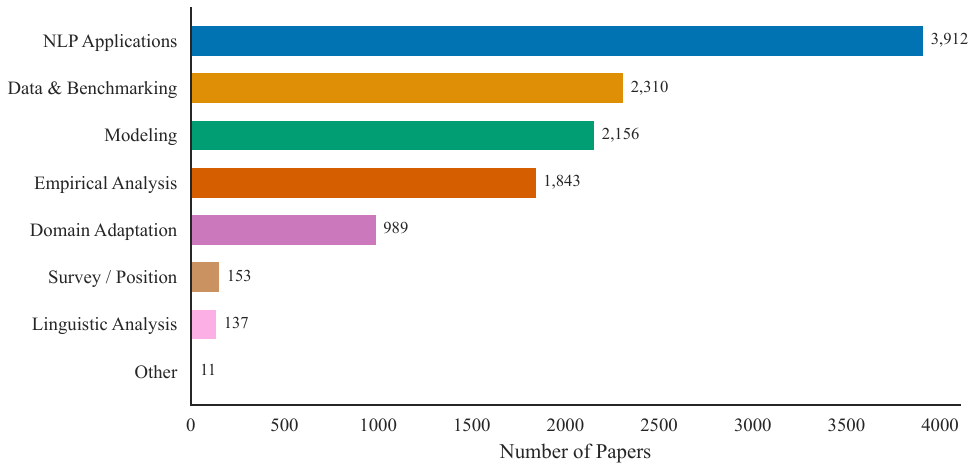}
  \caption{Distribution of contribution types in the full analysis corpus (Table~\ref{tab:analysis-corpus}), inferred from paper titles and abstracts. A paper may have multiple contribution types.}
  \label{fig:contribution-distribution}
\end{figure*}

Figure~\ref{fig:contribution-distribution} shows the distribution of contribution types across the full analysis corpus, similarly inferred from paper titles and abstracts using the contribution type prompt (see Appendix~\ref{app:prompts}). As with languages, a single paper may be assigned multiple contribution types; the counts reflect the number of papers associated with each category.

\section{Bias Rate Breakdowns by Venue Year and Contribution Type}
\label{app:bias-figures}

\begin{figure}
    \centering
    \includegraphics[width=0.8\linewidth]{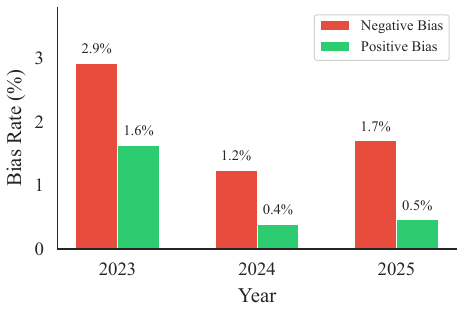}
    \caption{Negative vs.\ positive bias rate by venue year.}
    \label{fig:bias-type-by-year}
  \end{figure}

Figure~\ref{fig:bias-type-by-year} breaks down predicted bias rates by venue year. Negative bias rates are highest in 2023 and appear to decrease slightly in later cycles, while positive bias rates remain low throughout. However, the time span covers only three years (2023--2025), and venue composition differs across years (e.g., the 2024 data includes rejected ARR submissions absent from other years), so these trends should not be interpreted as evidence of a clear temporal trajectory. A longer observation window with controlled venue coverage would be needed to draw reliable conclusions about temporal change.

  \begin{figure*}
    \centering
    \includegraphics[width=0.7\linewidth]{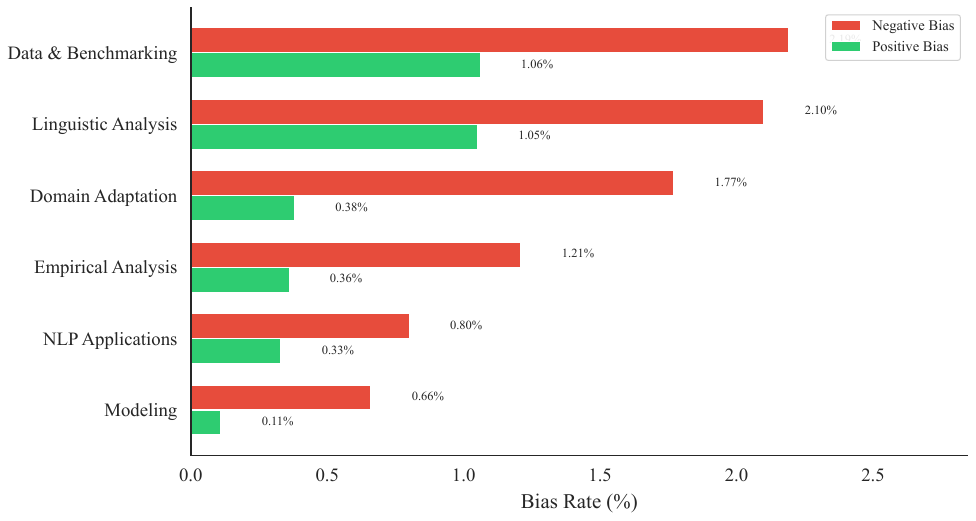}
    \caption{Review-level negative vs.\ positive bias rate by contribution type. Survey/Position and Other categories are excluded (zero bias instances).}
    \label{fig:bias-type-by-contrib}
  \end{figure*}

\begin{figure*}
  \centering
  \includegraphics[width=0.75\linewidth]{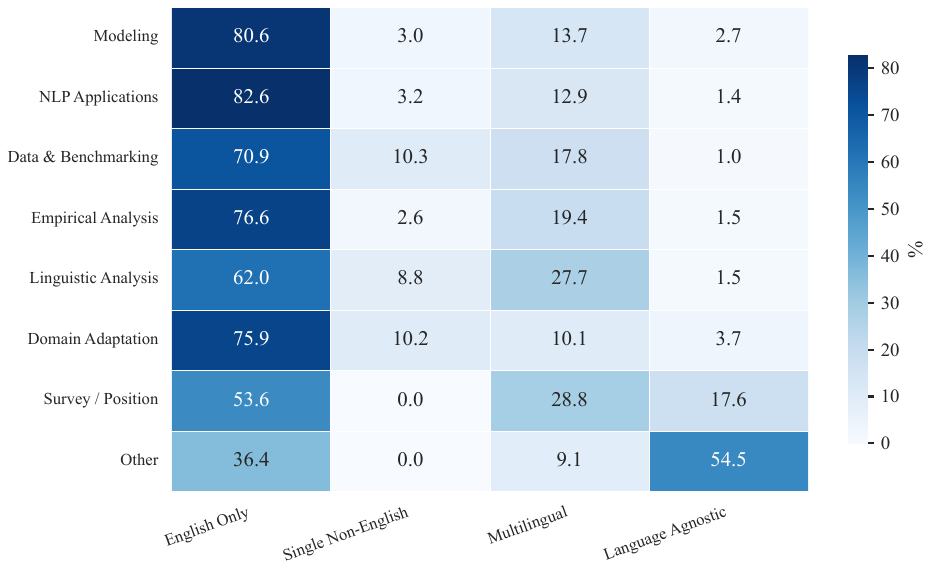}
  \caption{Cross-tabulation of studied language and contribution type, showing the number of papers at each intersection. Only languages with $\geq 5$ papers are shown.}
  \label{fig:language-x-contribution}
\end{figure*}

\begin{figure*}
  \centering
  \includegraphics[width=0.75\linewidth]{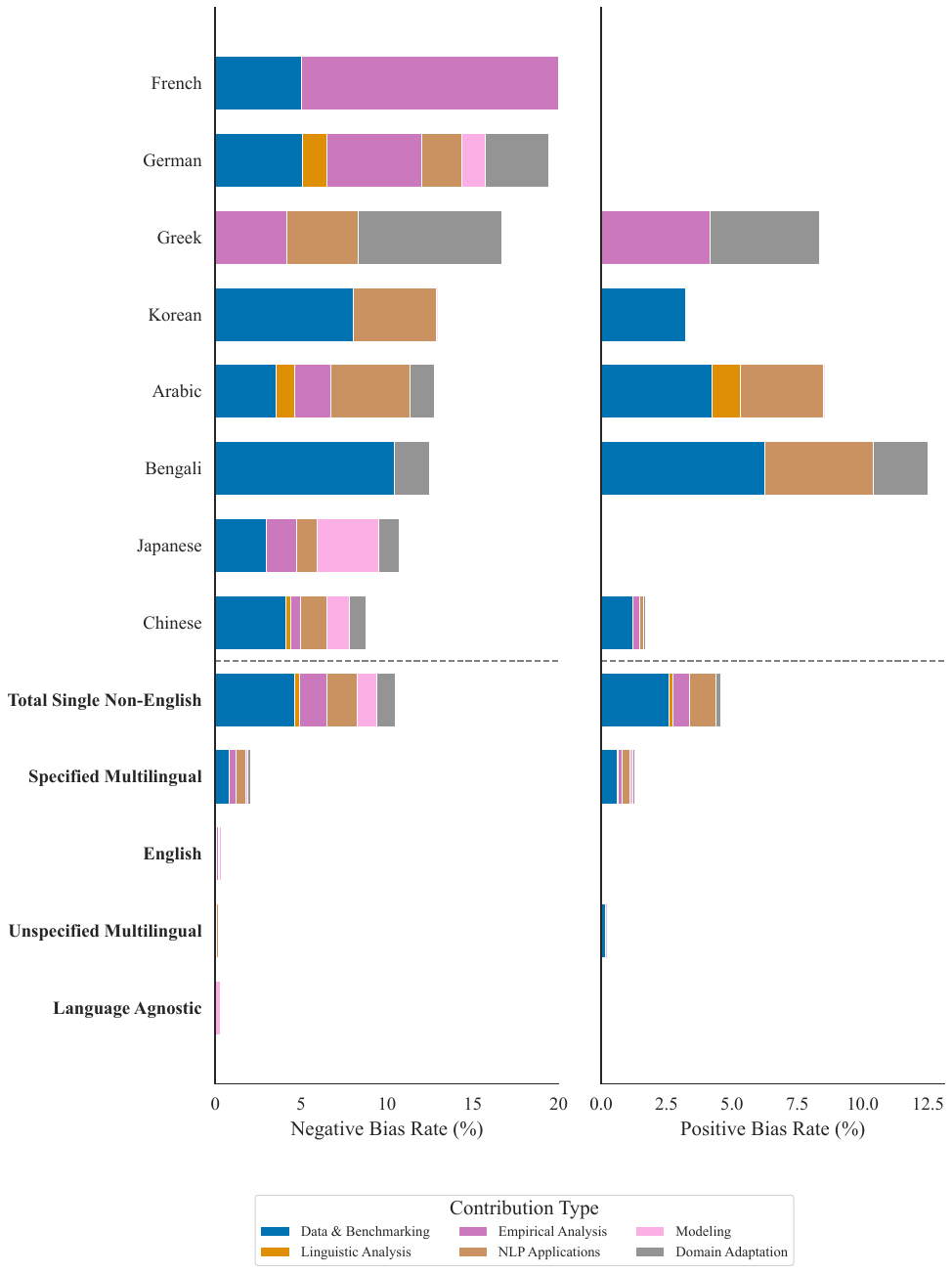}
  \caption{Decomposition of per-language negative and positive bias rates by paper contribution type.}
  \label{fig:bias-decomposition-contrib-neg-pos}
\end{figure*}

\section{Context Ablation}
\label{app:context-ablation}

\revised{Our bias detector receives the paper title, abstract, and the full review text (not a single comment) at inference time. To probe how much paper-scope context matters, we re-run \textsc{Gemini-3.1-Pro-Preview} on \dataset{} under the settings in Table~\ref{tab:context-ablation}. Removing the title and abstract costs $4.21$ Macro F1. We hypothesize that, the drop is relatively low since reviews themselves embed scope information. Reviewer written summary fields (e.g., the ARR \emph{Paper Summary}; the EMNLP~2023 \emph{Paper Topic and Main Contributions}) restate the paper's scope, and evaluative comments reference it as well (e.g., listing a new Korean ASR benchmark as a strength). Consistent with the latter, stripping the summary fields as well leaves performance essentially unchanged ($84.81$ vs.\ $83.16$). Replacing the title and abstract with those of a randomly assigned \emph{different} paper reduces the performance by $4.17$ points, comparable to removing them outright. %
Furthermore, we find that adding the full paper text hurts the performance: on the 525-segment subset for which we can retrieve PDFs (439 papers across EMNLP~2023 via OpenReview, ACL~2025 via the ARR Data Collection Initiative, and EMNLP~2024 via ACL Anthology camera-ready PDFs), Macro F1 drops from $87.36$ to $80.27$. The same direction holds within each venue (EMNLP~2023: $89.16 \rightarrow 82.39$; EMNLP~2024: $78.34 \rightarrow 72.35$; ACL~2025: $60.15 \rightarrow 52.10$, on only 56 segments and therefore noisy). We believe this is due to the full-PDF context burying the bias-relevant signal rather than amplifying it, suggesting that most evidence for language-of-study bias lives in the reviewer's own text. The asymmetry between annotator and detector context (\S\ref{ssec:annotation}) is therefore unlikely to be a substantial source of error, and the title+abstract+review setting we use throughout the corpus is a reasonable trade-off between coverage and detector signal.}

\begin{table}[t]
\centering
\small
\begin{tabular}{lrr}
\toprule
\textbf{Context given to the detector} & \textbf{\llap{Macro} F1} & \textbf{$\Delta$} \\
\midrule
\multicolumn{3}{@{}l}{\emph{Full \dataset{} (529 segments)}} \\
Title, abstract, review (default)      & 87.37 & --      \\
Review only                            & 83.16 & $-4.21$ \\
Review without reviewer summaries      & 84.81 & $-2.56$ \\
Random paper's title, abstract, review & 83.20 & $-4.17$ \\
\addlinespace
\multicolumn{3}{@{}l}{\emph{Subset with retrievable PDFs (525 segments)}} \\
Title, abstract, review (default)      & 87.36 & --      \\
\quad $+$ full paper text              & 80.27 & $-7.09$ \\
\bottomrule
\end{tabular}
\caption{\revised{Detector Macro F1 on \dataset{} under different context settings; $\Delta$ is relative to the default setting of the same panel. Reviews typically contain a reviewer-written summary of the paper, so the third row strips those fields as well.}}
\label{tab:context-ablation}
\end{table}

\section{Analysis of Detector Errors and Annotator Disagreement}
\label{app:annotator-analyses}
\revised{We analyze the errors of our bias detector, annotator disagreement and how they compare. Of the 529 segments, 89 (17\%) received a split vote: 68 were 2:1 splits among three annotators and 21 were splits among five. For each segment we compute the Shannon entropy~\cite{shannon1948} of the annotator vote distribution (zero when annotators agree, higher when votes split). Mean entropy is $0.21$ for misclassified segments versus $0.15$ for correct ones (a 1.4$\times$ ratio); the difference is in the expected direction although not statistically significant on this small set ($n_{\text{err}} = 34$, Mann--Whitney $p = 0.14$). This is consistent with detector errors concentrating on segments that even trained annotators found ambiguous. We find three common cases that cause ambiguity: First, reviewers often ask for additional languages alongside domains, datasets, or model sizes, and annotators differed on whether this is a standard generalizability critique or a multilingual requirement imposed on a paper that never claimed cross-lingual generality. A second group involves English as a default baseline, pivot language, or validation standard, which can be read either as a concrete methodological suggestion (e.g., a translate-test baseline) or as privileging English without justification. Disagreement also arose when a paper's own scope was unclear: if it proposes a ``general method'' but evaluates a single language, a multilingual critique is hard to classify as either scope-aligned or biased. In each case, the label depends on how tightly reviewer expectations are anchored to the paper's stated scope.}

\section{Annotation Prompts}
\label{app:prompts}

See the last pages of the appendix for the prompts used for annotating the
\textit{language-of-study bias}
(all inferences are performed with temperature $=0.0$, top-$p=0.95$, and random seed $42$ to ensure reproducibility),
the \textit{language-of-study},
the \textit{contribution type},
and the \textit{bias subcategory}.

\clearpage
\onecolumn

\begin{promptwindow}{Bias Toward/Against Studied Languages Prompt}
  \small

  \vspace{3pt}
  \textbf{Task}

  You are an expert in NLP peer review analysis. Identify \textbf{language bias} — cases where a reviewer evaluates a paper differently because of which natural language(s) it studies, rather than on scientific merit.

  \vspace{3pt}
  \textbf{Language} = human natural languages, dialects, varieties, and sign languages. Excludes programming languages and constructed languages (Esperanto, Klingon).

  \vspace{3pt}
  \textbf{This task is NOT about} cultural/geographic topic scope (e.g., "American political context"), nor general domain/topic niche-ness unless explicitly tied to the language(s) studied.

  \vspace{3pt}
  \textbf{Decision Process}

  In your internal reasoning, first read the title and abstract carefully, then analyze the full review in context before making any judgment. For each reviewer comment about language scope, apply these two tests in order:

  \vspace{3pt}
  \textbf{Step 1 — Paper Scope.} Read the title and abstract to determine the paper's language scope:

  \begin{itemize}
    \item \textbf{Scoped}: Paper is explicitly limited to specific language(s) and makes no claims of multilingual, cross-lingual, or language-independent generalizability.
    \item \textbf{Claims generalizability}: Paper claims broad applicability ("language-agnostic", "multilingual", "cross-lingual", etc.) or does not clearly limit its scope.
  \end{itemize}

  If the paper \textbf{claims generalizability} or \textbf{does not limit its scope} $\rightarrow$ reviewer comments about other languages or cross-lingual performance are \textbf{valid scientific feedback}, not bias. Stop here.

  If the paper is \textbf{scoped} $\rightarrow$ proceed to Step 2.

  \vspace{3pt}
  \textbf{Step 2 — Review Section \& Impact.} Where does the comment appear, and does it affect the reviewer's decision?

  \begin{itemize}
    \item In \textbf{Weaknesses / Reasons to Reject / Major Concerns}, or used to justify a low score $\rightarrow$ \textbf{strong bias signal}. Read the surrounding context carefully to confirm bias before flagging.
    \item In \textbf{Questions / Suggestions / Future Work} only, without being tied to the rejection decision or downgrading the current contribution $\rightarrow$ \textbf{not bias}. Do not flag.
    \item In \textbf{Strengths / Reasons to Accept}, as the primary/sole justification for acceptance $\rightarrow$ \textbf{positive bias signal}. Read the surrounding context carefully to confirm bias before flagging.
  \end{itemize}

  \vspace{3pt}
  \textbf{Two Types of Language Bias}

  \vspace{3pt}
  \textbf{Negative Bias – Flag if the reviewer:}

  \begin{itemize}
    \item Treats English (or another high-resource language) as a required validity check, even though the paper is scoped to non-English language(s).
    \item Insists the work is incomplete or unconvincing without English evaluation or popular English benchmarks (e.g., GLUE, SuperGLUE).
    \item Demands multilingual or cross-lingual experiments as a prerequisite for acceptance when the paper never claims generalizability.
    \item Questions the paper's applicability/generalizability primarily because experiments are on a single language, when the paper is clearly scoped.
    \item Downplays the paper's impact, relevance, or venue fit because it focuses on a low-resource, non-English, or lesser-studied language (e.g., "few researchers will care", "too niche", "limited audience", "better suited for a workshop").
    \item States the contribution is small/weak because the language(s) are perceived as minor, obscure, or not "major."
    \item Questions or rejects the motivation for the language choice as if the language is unworthy of study (e.g., "why this obscure language?", "why not a more widely studied language?").
    \item Frames limited language scope as a critical flaw when broad generalization was never claimed.
  \end{itemize}

  \vspace{3pt}
  \textbf{Positive Bias – Flag if the reviewer:}

  Positive bias should be \textbf{rare}. The core pattern is: the reviewer praises the paper's language choice as a merit \textbf{in itself}, disconnected from the paper's actual methods, results, or novelty.

  \begin{itemize}
    \item The language choice appears as a \textbf{standalone reason} in Strengths/Accept, not connected to specific methods or results (e.g., "The creation of a new dataset in a non-English language.", "The research here is on the Chinese language.").
    \item The praise is \textbf{generic and unconditional} — it could apply to any paper in that language. Look for: "always valuable", "in itself", "the real contribution is the dataset for [language]."
    \item The language is framed as the \textbf{primary justification} for the paper's value rather than its methodology or findings.
  \end{itemize}

  \vspace{3pt}
  \textbf{Do NOT Flag (Valid Critique)}

  \begin{itemize}
    \item Criticism of dataset size, data quality, annotation process, baselines, ablations, reproducibility, etc. — regardless of the language studied.
    \item Requests for additional language experiments when the paper \textbf{claims} language-independence, cross-lingual transfer, or broad applicability.
    \item Suggestions phrased as optional improvements ("it would be nice to see…", "future work could…") without affecting the accept/reject decision.
    \item Requests for methodological rigor (e.g., asking why a particular baseline is used) without implying the language is unworthy.
    \item Comments about accessibility (e.g., adding English translations for readability) framed as optional improvements.
  \end{itemize}

  \vspace{3pt}
  \textbf{Annotation Rules}

  \begin{itemize}
    \item Quote the \textbf{full sentence} containing the biased statement (not just a clause).
    \item If a sentence mixes bias with valid critique, still quote the \textbf{entire sentence}.
    \item Output every distinct biased claim separately, even if from the same reviewer.
    \item Annotate bias in the reviewer text only, never in the paper itself.
  \end{itemize}

  \vspace{3pt}
  \textbf{Output Format (JSON)}

  Return \textbf{only} valid JSON. No markdown fences, no extra commentary. If no bias: \texttt{{"biases": []}}.

  \begin{verbatim}
{
  "biases": [
    {
      "quoted_text": "<biased statement>",
      "type": "<negative | positive>",
      "justification": "<1-3 sentences
        explaining why this is bias>"
    }
  ]
}
\end{verbatim}

  \vspace{3pt}
  \textbf{Input Data}

  \vspace{3pt}
  \textbf{Paper Title:} \texttt{\{title\}}

  \vspace{3pt}
  \textbf{Abstract:}
  \texttt{\{abstract\}}

  \vspace{3pt}
  \textbf{Review Text:}
  \texttt{\{review\_text\}}

\end{promptwindow}

\vspace{10pt}

\begin{promptwindow}{Languages Studied Identification Prompt}
  \small

  \vspace{3pt}
  \textbf{Role}

  You are an expert annotator of NLP papers.

  \vspace{3pt}
  \textbf{Task}

  Your task is to determine the \textbf{natural languages} studied in a given paper based on the Title, Abstract, and Reviews.

  \vspace{3pt}
  \textbf{Steps}

  \begin{enumerate}
    \item \textbf{Analyze Evidence:} Look for specific mentions of languages, datasets (infer the language if the dataset is standard, e.g., SQuAD = English), and claims in the abstract/reviews.
    \item \textbf{Filter:} Exclude programming languages (Python, Java, etc.) unless the task involves natural-language-to-code translation.
    \item \textbf{Synthesize:} Write reasoning into the "justification" field.
    \item \textbf{Output:} Generate a valid JSON object.
  \end{enumerate}

  \vspace{3pt}
  \textbf{Annotation Rules}

  \vspace{3pt}
  \textbf{1. Naming \& Normalization}

  \begin{itemize}
    \item \textbf{Explicit Mentions:} Output the full English name (no ISO codes).
          \begin{itemize}
            \item Bad: "en", "de", "MSA"
            \item Good: "English", "German", "Arabic"
            \item \textbf{Normalization Map:}
            \item "Mandarin", "Putonghua", "Cantonese", "Taiwanese Mandarin", "Simplified Chinese", "Traditional Chinese" $\rightarrow$ "Chinese"
            \item "Modern Standard Arabic", "MSA", "Egyptian Arabic", "Gulf Arabic", "Levantine Arabic", "Maghrebi Arabic" $\rightarrow$ "Arabic"
            \item "Farsi" $\rightarrow$ "Persian"
            \item "Castilian" $\rightarrow$ "Spanish"
            \item "Swiss German", "Austrian German", "Bavarian" $\rightarrow$ "German"
            \item "Brazilian Portuguese", "European Portuguese" $\rightarrow$ "Portuguese"
            \item "Scots English", "Australian English", "Indian English" $\rightarrow$ "English"
            \item \textbf{Dialects:} Treat dialects as the parent language (e.g., "Cantonese" $\rightarrow$ "Chinese", "Swiss German" $\rightarrow$ "German").
            \item \textbf{Dialect-focused papers:} If the paper's research goal is specifically to study dialect differences (e.g., "A Multidialectal Dataset of Arabic Proverbs"), still normalize to the parent language (e.g., "Arabic") but note the dialect focus in the justification.
          \end{itemize}
  \end{itemize}

  \vspace{3pt}
  \textbf{2. Language Scope Categories}

  Classify each paper into exactly one \texttt{language\_scope} category:

  \begin{itemize}
    \item \texttt{single-language} --- One specific language studied.
    \item \texttt{multilingual-specified} --- Multiple specific languages listed.
    \item \texttt{multilingual-partial} --- Some languages named + ``others'' implied; list only the named ones.
    \item \texttt{multilingual-count-only} --- Only a count given (e.g., ``101 languages'').
    \item \texttt{multilingual-unspecified} --- Vague ``multilingual'' claim, no names/counts.
    \item \texttt{language-agnostic} --- No natural language involved.
  \end{itemize}

  \vspace{3pt}
  \textbf{3. Handling Counts}

  \begin{itemize}
    \item \texttt{languages\_count}: Number of unique languages in the \texttt{languages} list.
    \item For \texttt{multilingual-count-only}: Use the stated count (e.g., 101) even though \texttt{languages} is empty.
    \item For \texttt{multilingual-unspecified} and \texttt{language-agnostic}: Set to 0.
  \end{itemize}

  \vspace{3pt}
  \textbf{4. Defaults \& Edge Cases}

  \begin{itemize}
    \item \textbf{English Default:} If datasets are known to be English (e.g., GLUE, SQuAD, ImageNet) and no other language is mentioned $\rightarrow$ \texttt{language\_scope}: \texttt{single-language}, \texttt{languages}: \texttt{["English"]}.
    \item \textbf{Language-Agnostic:} If the method is purely mathematical/symbolic or applied \textit{only} to synthetic data/pixels without text $\rightarrow$ \texttt{language\_scope}: \texttt{language-agnostic}, \texttt{languages}: \texttt{[]}.
    \item \textbf{Programming Languages:} Do not list "Python" or "C++" in \texttt{languages}. If the paper uses English prompts to generate Python code, the language studied is "English". If the code generation benchmark is multilingual (e.g., MultiPL-E), list the natural languages of the prompts/docstrings used.
    \item \textbf{Sign Languages:} Sign languages (e.g., American Sign Language, British Sign Language) are natural languages and should be included. Normalize to the specific sign language name (e.g., "American Sign Language"). Do not collapse different sign languages into one.
    \item \textbf{Romanized/Transliterated Text:} If a paper studies text in a romanized form (e.g., Hindi written in Latin script), the language is still the original language (e.g., "Hindi"), not "English".
  \end{itemize}

  \vspace{3pt}
  \textbf{5. Priority between evidence sources (title/abstract vs reviews)}

  \begin{itemize}
    \item Primary evidence = actual experiments and evaluations described (first in abstract, then in reviews if abstract is vague or missing details).
    \item If reviews explicitly describe the evaluated languages (e.g., “They only evaluate on English,” “Experiments are English-only,” “No non-English results reported”), trust the reviews over broad claims in the abstract/title.
    \item Ignore speculative reviewer suggestions (e.g., “They should evaluate on Chinese”)—only count what was actually done.
    \item \textbf{Training vs. Evaluation languages:} Focus on languages used in \textbf{evaluation/testing}. If a model is trained on English but evaluated on Hindi and Chinese, the languages are \texttt{["Hindi", "Chinese"]}. If both training and evaluation languages are explicitly part of the study's contribution, include all of them.
  \end{itemize}

  \vspace{3pt}
  \textbf{6. Evidence type priority (choose exactly one)}

  Use the highest-priority category that applies:

  \begin{enumerate}
    \item \textbf{explicit\_list} — any specific natural language names are mentioned as being experimentally evaluated (highest priority; overrides everything else).
    \item \textbf{dataset\_implied} — no explicit language names, but languages can be reliably inferred from standard dataset names.
    \item \textbf{count\_only} — only a number of languages is given (e.g., "101 languages") without names or identifiable datasets.
    \item \textbf{claim\_only} — only vague claims like "multilingual" or "cross-lingual" with no names, datasets, or counts (lowest priority).
  \end{enumerate}

  \vspace{3pt}
  \textbf{Output Fields}

  \begin{itemize}
    \item \texttt{language\_scope}: One of: \texttt{"single-language"}, \texttt{"multilingual-specified"}, \texttt{"multilingual-partial"}, \texttt{"multilingual-count-only"}, \texttt{"multilingual-unspecified"}, \texttt{"language-agnostic"}.
    \item \texttt{languages}: Array of normalized language names (can be empty).
    \item \texttt{languages\_count}: Integer (length of \texttt{languages} array, or stated count for \texttt{multilingual-count-only}).
    \item \texttt{evidence\_type}: \texttt{"explicit\_list"}, \texttt{"dataset\_implied"}, \texttt{"count\_only"}, or \texttt{"claim\_only"}.
    \item \texttt{justification}: A concise string citing the specific text snippets or dataset names that led to the decision.
  \end{itemize}

  \vspace{3pt}
  \textbf{Output Format (JSON)}

  \vspace{3pt}
  \textbf{Example 1: Single language (inferred from dataset)}

  \begin{verbatim}
{
  "language_scope": "single-language",
  "languages": ["English"],
  "languages_count": 1,
  "evidence_type": "dataset_implied",
  "justification": "The abstract mentions
    benchmarking on VSR and A-OKVQA, which
    are standard English-language datasets."
}
\end{verbatim}

  \vspace{3pt}
  \textbf{Example 2: Multilingual with specific languages}

  \begin{verbatim}
{
  "language_scope": "multilingual-specified",
  "languages": [
    "Bhojpuri",
    "Hindi",
    "Meadow Mari",
    "Russian",
    "Tibetan",
    "English"
  ],
  "languages_count": 6,
  "evidence_type": "explicit_list",
  "justification": "The abstract explicitly
    lists experiments on Bhojpuri, Hindi,
    Meadow Mari, Russian, Tibetan,
    and English."
}
\end{verbatim}

  \vspace{3pt}
  \textbf{Example 3: Multilingual with count only}

  \begin{verbatim}
{
  "language_scope": "multilingual-count-only",
  "languages": [],
  "languages_count": 101,
  "evidence_type": "count_only",
  "justification": "The abstract states
    'We evaluate on 101 languages' but does
    not list specific language names."
}
\end{verbatim}

  \vspace{3pt}
  \textbf{Example 4: Language-agnostic}

  \begin{verbatim}
{
  "language_scope": "language-agnostic",
  "languages": [],
  "languages_count": 0,
  "evidence_type": "claim_only",
  "justification": "The paper proposes a
    mathematical optimization method for
    neural architecture search with no
    text data involved."
}
\end{verbatim}

  \vspace{3pt}
  \textbf{Input Data}

  \vspace{3pt}
  \textbf{Paper Title:} \texttt{\{title\}}

  \vspace{3pt}
  \textbf{Abstract:}
  \texttt{\{abstract\}}

  \vspace{3pt}
  \textbf{Reviews:}
  \texttt{\{reviews\_text\}}

\end{promptwindow}

\begin{promptwindow}{Contribution Type Classification Prompt}
  \small

  \vspace{3pt}
  \textbf{Role}

  You are an expert annotator of NLP research papers.

  \vspace{3pt}
  \textbf{Task}

  Using ONLY the paper title and abstract, identify the paper's main contribution type(s),
  following the categories below.

  Select ONE OR MORE labels from the list below.

  \begin{itemize}
    \item Use a label only if it is clearly supported by the title or abstract.
    \item If multiple contribution types are present, include all that apply.
    \item If none clearly apply, select \texttt{Other}.
    \item Do not infer beyond the given text.
  \end{itemize}

  \vspace{3pt}
  \textbf{Contribution Types (use EXACT strings)}

  \textbf{\texttt{Modeling}}

  Proposes a new model, architecture, learning algorithm, objective function, or decoding method.

  \begin{itemize}
    \item \textbf{Use when} the abstract emphasizes a novel \textit{technical component} — something that changes how a model is structured, trained, or performs inference (e.g., attention mechanism, loss function, pre-training objective, model compression technique).
    \item \textbf{Do NOT use} for papers that merely fine-tune or apply an existing model without architectural or algorithmic novelty.
  \end{itemize}

  \textbf{\texttt{NLPApplications}}

  Introduces a new pipeline, system, prompting strategy, data augmentation technique, or training procedure built on top of existing models.

  \begin{itemize}
    \item \textbf{Use when} the core novelty is a \textit{workflow, system integration, or engineering strategy} rather than a new model architecture. Includes retrieval-augmented generation pipelines, multi-agent systems, and prompt engineering methods.
    \item \textbf{Distinguish from Modeling:} if the contribution could work with different underlying models, it is \texttt{NLPApplications}; if it changes the model itself, it is \texttt{Modeling}.
  \end{itemize}

  \textbf{\texttt{DataAndBenchmarking}}

  Creates a new dataset, corpus, treebank, lexicon, knowledge base, annotation resource, benchmark, evaluation suite, shared task, or defines a new NLP task with accompanying data.

  \begin{itemize}
    \item \textbf{Use when} the abstract foregrounds the \textit{artifact itself} — its construction, curation, or novelty — as the primary contribution.
  \end{itemize}

  \textbf{\texttt{EmpiricalAnalysis}}

  Conducts a systematic empirical study of existing models or methods — such as comparative evaluations, ablation studies, error analyses, reproducibility/replication studies, or interpretability investigations.

  \begin{itemize}
    \item \textbf{Use when} the abstract focuses on \textit{measuring, comparing, or explaining} existing systems.
    \item \textbf{Do NOT use} when the empirical study is secondary to a new model or resource.
    \item \textbf{Do NOT use} when the analysis centers on a linguistic phenomenon rather than model behavior — use \texttt{LinguisticAnalysis} instead.
  \end{itemize}

  \textbf{\texttt{LinguisticAnalysis}}

  Investigates a specific linguistic phenomenon, typological property, psycholinguistic question, or language-specific feature using computational methods (e.g., morphological analysis, code-switching patterns, dialectal variation, cross-lingual transfer properties, reading behavior, language acquisition, cognitive processing of language).

  \begin{itemize}
    \item \textbf{Use when} the primary goal is advancing \textit{understanding of language or its cognitive processing}.
    \item \textbf{Do NOT use} for papers that simply test a model on language-specific data without linguistic analysis.
    \item \textbf{Distinguish from EmpiricalAnalysis:} if the paper asks "how does model X perform?" it is \texttt{EmpiricalAnalysis}; if it asks "how does language phenomenon Y work?" it is \texttt{LinguisticAnalysis}.
  \end{itemize}

  \textbf{\texttt{DomainAdaptation}}

  Applies or adapts NLP techniques to a specific real-world domain (e.g., clinical NLP, legal text processing, scientific document analysis, educational technology, social media analysis).

  \begin{itemize}
    \item \textbf{Use when} the abstract emphasizes the \textit{domain context and domain-specific challenges or insights} as the contribution, rather than general-purpose NLP methodology.
  \end{itemize}

  \textbf{\texttt{SurveyOrPosition}}

  Provides a structured literature review, meta-analysis, systematic mapping study, or argues a conceptual position or research agenda.

  \begin{itemize}
    \item \textbf{Use when} the primary contribution is \textit{synthesis of prior work or argumentation}, not new empirical results. Includes tutorial-style overview papers.
  \end{itemize}

  \textbf{\texttt{Other}}

  Does not clearly fit any of the above categories.

  \vspace{3pt}
  \textbf{Output Format (JSON only)}

  Important: Reason through the classification before selecting labels.

  \begin{verbatim}
{
  "justification": "A brief (1-2 sentence)
    explanation of why the selected labels
    apply, citing specific keywords or
    claims from the abstract.",
  "contribution_type": ["Label1", "Label2"]
}
\end{verbatim}

  \vspace{3pt}
  \textbf{Input}

  Paper Title: \texttt{\{title\}}

  Abstract:
  \texttt{\{abstract\}}

\end{promptwindow}

\vspace{\baselineskip}

\begin{promptwindow}{Negative Bias Subcategory Classification Prompt}
  \small

  \vspace{3pt}
  \textbf{Task}

  You are an expert in NLP peer review analysis. Your task is to classify a peer review that has already been identified as containing language bias into one of the following four patterns.

  \vspace{3pt}
  Language bias occurs when a reviewer penalizes a paper because of which natural language(s) it studies, rather than on scientific merit. This covers human natural languages, dialects, and varieties only — not programming languages or topic/cultural scope unless explicitly tied to the language studied.

  \vspace{3pt}
  \textbf{Four Patterns}

  \vspace{3pt}
  \textbf{A — Generalizability Demand}

  The reviewer penalizes the paper for not demonstrating cross-lingual generalizability that was never claimed, treating multilingual coverage as a precondition for validity.

  Signal: language scope in Weaknesses/Reject despite no generalizability claims in paper.

  \vspace{3pt}
  \textbf{B — English as the Gold Standard}

  The reviewer explicitly names English as the missing validation standard, implying non-English results are insufficient without English corroboration.

  Signal: English named specifically (not just ``other languages'') as a required benchmark.

  \vspace{3pt}
  \textbf{C — Language Choice Interrogation}

  The reviewer questions why a specific language was chosen, treating the language selection itself as requiring special justification — a standard not applied to English or high-resource languages.

  Signal: reviewer asks ``why X?'' or suggests alternative languages as more worthy choices.

  \vspace{3pt}
  \textbf{D — Impact Discounting}

  The reviewer accepts the paper's validity but diminishes its significance because the language community it serves is too small, or because adapting to a new language is not considered genuine novelty.

  Signal: ``only useful to X community,'' novelty denied on language grounds, workshop suggestion, contradiction between praising the work and penalizing its audience size.

  \vspace{3pt}
  \textbf{Important}

  Only classify as bias if the language scope concern functions as a penalty — i.e., it appears in Weaknesses or Reasons to Reject and is not solely a neutral suggestion. If the paper itself claims cross-lingual generalizability, questioning that claim is legitimate, not bias.

  \vspace{3pt}
  \textbf{Output Format (JSON)}

  Respond in \textbf{only} valid JSON. No text outside the JSON block. Identify the single most prominent pattern in the review.

  \begin{verbatim}
{
  "evidence": "exact quote from review
    that best represents the bias",
  "pattern": "A",
  "reasoning": "What the reviewer did +
    why it constitutes bias of
    this pattern."
}
\end{verbatim}

  \vspace{3pt}
  \textbf{Input Data}

  \vspace{3pt}
  \textbf{Paper Title:} \texttt{\{title\}}

  \vspace{3pt}
  \textbf{Abstract:}
  \texttt{\{abstract\}}

  \vspace{3pt}
  \textbf{Review:}
  \texttt{\{review\}}

\end{promptwindow}

\end{document}